\documentclass[sigconf]{acmart} 
\AtBeginDocument{%
}

\setcopyright{acmcopyright}
\copyrightyear{2023}
\acmYear{2023}
\acmDOI{XXXXXXX.XXXXXXX}

\acmConference[MM '23]{Proceedings of the 31st ACM International Conference on Multimedia}{October 28, 2023 -- November 3, 2023}{Ottawa, Canada}
\acmBooktitle{Proceedings of the 31st ACM International Conference on Multimedia (MM '23), October 28, 2023 -- November 3, 2023, Ottawa, Canada}

\acmPrice{15.00}
\acmISBN{978-1-4503-XXXX-X/18/06}

\acmSubmissionID{1558}

\usepackage{algorithm}
\usepackage{algorithmic}
\usepackage{color}
\usepackage{bm}
\usepackage{multirow}
\usepackage[american]{babel}
\usepackage{microtype}
\newcommand{\T}{{\top}}
\newcommand{\z}{{\bm{z}}}

\newcommand{\x}{{\bm{x}}}

\newcommand{\eg}{\textrm{e.g.}}
\newcommand{\ie}{\textrm{i.e.}}
\newcommand{\wrt}{\textit{w.r.t.\,}}
\newcommand{\blue}{\textcolor[rgb]{0,0,0}}
\newcommand{\red}{\textcolor[rgb]{0,0,0}}

\begin{document}

\title{Entropy Neural Estimation for Graph Contrastive Learning}

\author{Yixuan Ma}
\email{mayx2021@lzu.edu.cn}
\affiliation{%
  \institution{Lanzhou University}
  \city{Lanzhou}
  \country{China}
}

	\author{Xiaolin Zhang}
\email{solli.zhang@gmail.com}
\affiliation{%
	\institution{}
	\city{Shenzhen}
	\country{China}
}

\author{Peng Zhang}
\email{pengzhang_skd@sdust.edu.cn}
\affiliation{%
	\institution{College of Computer Science and Engineering, \\
		Shandong University of Science and Technology}
	\city{Qingdao}
	\country{China}
}

\author{Kun Zhan}
\authornote{Corresponding author.}
\email{kzhan@lzu.edu.cn}	
\affiliation{%
	\institution{School of Information Science and Engineering, \\
		Lanzhou University}
	\city{Lanzhou}
	\country{China}
}

\renewcommand{\shortauthors}{Ma et al.}

\begin{abstract}
  Contrastive learning on graphs aims at extracting distinguishable high-level representations of nodes. In this paper, we theoretically illustrate that the entropy of a dataset can be approximated by maximizing the lower bound of the mutual information across different views of a graph, \ie, entropy is estimated by a neural network. Based on this finding, we propose a simple yet effective subset sampling strategy to contrast pairwise representations between views of a dataset. In particular, we randomly sample nodes and edges from a given graph to build the input subset for a view. Two views are fed into a parameter-shared Siamese network to extract the high-dimensional embeddings and estimate the information entropy of the entire graph. For the learning process, we propose to optimize the network using two objectives, simultaneously. Concretely, the input of the contrastive loss function consists of positive and negative pairs. Our selection strategy of pairs is different from previous works and we present a novel strategy to enhance the representation ability of the graph encoder by selecting nodes based on cross-view similarities. We enrich the diversity of the positive and negative pairs by selecting highly similar samples and totally different data with the guidance of cross-view similarity scores, respectively. We also introduce a cross-view consistency constraint on the representations generated from the different views. This objective guarantees the learned representations are consistent across views from the perspective of the entire graph. We conduct extensive experiments on seven graph benchmarks, and the proposed approach achieves competitive performance compared to the current state-of-the-art methods. The source code will be publicly released once this paper is accepted.
\end{abstract}



\maketitle

\section{Introduction}
The goal of unsupervised contrastive learning is to learn a representation of data such that similar instances are close together in the representation space, while dissimilar instances are far apart. This is achieved by maximizing an estimate of the mutual information between different views of the data.
Mutual information is a measure of the cross-dependence between two random variables and can be used to quantify how much information one variable contains about another. In the context of unsupervised contrastive learning, maximizing mutual information between different views of the data means that the algorithm tries to optimize the cross-dependency between these views in high-dimensional space. This can help improve the performance of models by allowing them to extract useful information from data.



Unsupervised contrastive learning is usually expressed as maximizing the mutual information between augmented multi-view data~\cite{oord2018representation,belghazi2018mutual}.
Given an unsupervised deep graph representation learning task,
the estimation process of the mutual information is to optimize the cross-dependency between different views of the data in high-dimensional space~\cite{belghazi2018mutual,poole2019variational}.
Practically, a lower bound of the information can be estimated by the contrastive learning methodology.

\begin{figure}[!t]
  \centering
  \includegraphics[width=\linewidth]{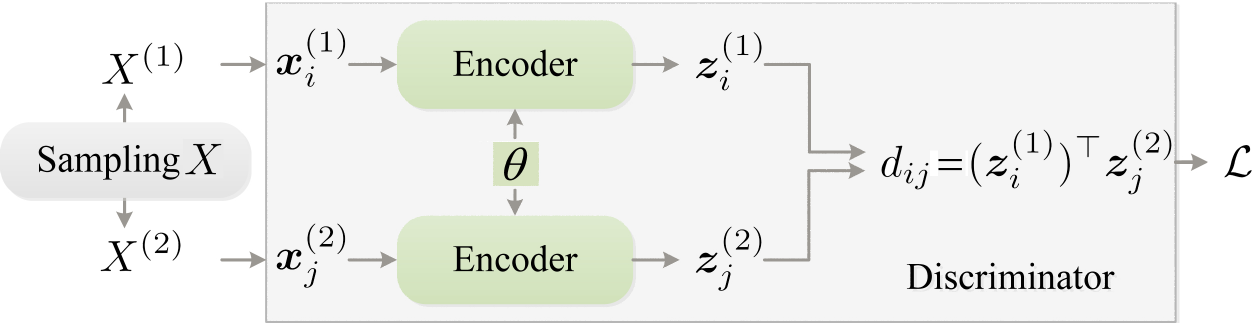}\\
  \caption{The architecture of the discriminator of $\mathcal{M}$-ILBO. Two-view subsets, $X^{(1)}$ and $X^{(2)}$, are sampled from the raw dataset $X$ in each epoch. Pairwise data points, $\x_i^{(1)}$ and $\x^{(2)}_j$, are inputs of the encoder and their similarity score is $d_{ij}\,$.  }\label{scheme}
\end{figure}

We propose to simulate view augmentation via subset sampling from raw data
and contrastively maximize the mutual information across subsets.
The training process can be viewed as the learning  of the correlations between two subsets of the given graph data.
As illustrated in Fig.~\ref{scheme}, we randomly sample two subsets of the input graph,~\ie,node features and graph edges.
Then, both of the subsets are fed into a shared graph encoder for extracting high-level representations of the input nodes .
Since both subsets are drawn from the same input graph, and seen as two different observation views of the input, the corresponded representations ought to consistent across them.
Therefore, we employ a discriminator to maintain the cross-view consistency for both subsets, where representations are forced to maintain close among high-similar nodes while stay apart between dissimilar nodes.
Constrastive learning maximizes the similarity between positive pairs of the same nodes from different views while minimize the similarity across negative pairs of different nodes.
Such a way of selecting positive and negative pairs violates the purpose of clustering features of the similar nodes.   
Differently, we think that the nodes with highly similar representations should be strengthed and treated as positive pairs, while only the nodes with totally dissimilar representations  can be used as negative pairs.
Thus, proposed a simple yet effective standards for selecting the positive and negative pairs to enrich the diversity of constructed self-supervisions. 
Besides the proposed method, we theoretically analyze that this approach can guarantee the maximization of the Information-entropy Lower BOund (ILBO) for a given dataset.
In particular, we use the Jensen-Shannon divergence parameterized by a neural network to maximize ILBO, \ie, we estimate entropy by maximizing ILBO.

\red{The contrastive loss in each epoch comprehensively maximizes ILBO, which is similar to Bagging algorithms, i.e., bootstrapping enhances diversity and the model learns to output representation with good discriminative ability of different classes.}
Particularly, the contrastive learning process compacts these representations with similar information in the encoded high-dimensional manifold. In contrastive learning, positive and negative pairs are included in the contrastive loss to scatter categories. A classical data pair partition strategy is that positive pairs have the same indices in two representations while negative pairs are selected in the same way except by doing random shuffling for one of the representations~\cite{velickovic2019deep,hassani2020contrastive}. Then the pairwise similarity scores are used to compute the contrastive loss. However, a defect of the classical strategy is that samples belonging to the same category are often treated as negative pairs.
This defect results in low accuracy for downstream classification tasks.
In contrast to the classical strategy, we consider further enriching the supervision information of positive and negative pairs with the guidance of the similarity scores.
Particularly, we add more positive pairs in non-diagonal whose similarity scores are high, and negative pairs whose similarity scores are low. Our intuition is based on the fact that nodes with high similarity are more likely to belong to the same category, while nodes with low similarity tend to belong to different categories. Selecting samples of the same category as positive pairs and samples of different categories as negative pairs does help to render the learned representations to contain more task-relevant information.

Besides the aforementioned contrastive objective, we innovatively propose to introduce a cross-view consistency loss to align the two views in a global perspective. The consistency loss is based on error modeling, \ie, we hypothesize the error of cross-view embeddings is governed by the Gaussian distribution and the cross-view consistency loss is defined by taking the negative logarithm of likelihood of the error. In practice, such consistency is realized by positive pairs, \ie, a positive pair is two representations of the same nodes in different views, and has been verified to be quite useful for the unlabeled samples~\cite{sajjadi2016regularization,tarvainen2017mean,berthelot2019mixmatch,xie2020unsupervised,zhang2021flexmatch}.
This cross-view consistency is to ensure that the predicted representation of the entire graph should be consistent when feeding different subsets as input.
Concretely, we choose a simple yet effective approach to constrain the cross-view consistency by using the $\ell_2$ norm of embeddings across views.

In a nutshell, we propose a maximizing ILBO ($\mathcal{M}$-ILBO) algorithm, in which we present a novel contrastive loss to train a parameter-shared graph encoder by two sampled subsets of the raw dataset. We argue that the estimation of information entropy can be achieved by gradient descent over neural networks. Further, a cross-view consistency regularization renders different views to produce the aligned information. With the contrastive loss and the cross-view consistency, $\mathcal{M}$-ILBO gathers more and more task-relevant information. Extensive experiments under various conditions are conducted to verify the superiorities of $\mathcal{M}$-ILBO on seven graph datasets. Compared to existing state-of-the-art unsupervised representation learning algorithms, $\mathcal{M}$-ILBO achieves competitive performances and exhibits satisfactory generalization ability.

The main contributions are summarized as follows:
\begin{itemize}
\item We theoretically analyze the entropy of a dataset can be estimated by maximizing ILBO. Based on this, we propose an effective approach that samples subsets of a dataset and maximizes their mutual information so as to maximize ILBO.
\item We propose an efficient graph representation learning framework with contrastive learning. The proposed $\mathcal{M}$-ILBO framework train a graph encoder with a novel contrastive loss and a cross-view consistency regularization.
\item We design a novel positive and negative pairs selection strategy in order to improve the representation ability. We use the similarity matrix to select pairs upon their scores. Through data sampling and contrastive learning between positive and negative pairs, the representation continuously learns more and more task-relevant information.
\end{itemize}
\section{Entropy Neural Estimation}
\subsection{Problem formulation}
In contrastive learning, the optimization objective is usually formulated as the cross-view mutual information lower bound maximization problem \cite{oord2018representation,belghazi2018mutual}. This means that the model's performance is seriously determined by the view augmentation. Inspired by the concept of Bagging \cite{breiman1996bagging} and Dropout \cite{srivastava2014dropout}, this paper regards subset sampling as view augmentation and reformulates the contrastive learning as a cross-subset mutual information entropy lower bound maximization problem.

Given a certain dataset $X$, ${\rm E}(X)$ denotes its information entropy. Notably, ${\rm E}(X)$ measures $X$'s implicit data distribution structure. According to the information theory, it is easy to know that $X$'s entropy equals its mutual information, \ie, ${\rm E}(X)={\rm I}(X, X)$. This can be addressed by neural estimation of the cross-view mutual information in contrastive learning. Unfortunately, the solution is seriously limited by view augmentation. Instead of well-designed view augmentation, we propose to approximate the information entropy by subset re-sampling. Particularly, the two sampled subsets, \ie, $X^{(i)},\forall\,i\in\{1,2\}$, can be seen as two views of the same dataset $X$, where $i$ indicates the $i$-th view. Without loss of generality, we have ${\rm I}(X, X) \geq {\rm I}(X^{(1)}, X^{(2)})$. 
We approach ${\rm E}(X)$ by maximizing the mutual information of pairwise subsets of dataset $X$. Then, we have
\begin{align}
{\rm E}(X)={\rm I}(X, X) \geq {\rm I}(X^{(1)}, X^{(2)})\,.\label{entropy}
\end{align}
A neural estimator of the information entropy can be realized by the accumulation of the mutual information between these re-sampled subsets.

The mutual information, ${\rm I}(X^{(1)},X^{(2)})$, is defined by
\begin{align}\label{MI}
{\rm I}(X^{(1)},X^{(2)})
={\rm div}_{\rm kl}\bigl(p(\x^{(1)},\x^{(2)})\|p(\x^{(1)})p(\x^{(2)})\bigr)
\end{align}
where $\x^{(i)} \in X^{(i)}, \forall\,i\in\{1,2\}$ denote the samples in subset $X^{(i)}$, and ${\rm div}_{\rm kl}$ denotes the Kullback-Leibler divergence.

Following contrastive learning~\cite{oord2018representation,belghazi2018mutual}, the above optimization of the mutual information between two subsets can turn into a lower bound maximization problem. In practice, we use Jensen-Shannon divergence instead of the Kullback-Leibler divergence as \citet{nowozin2016f,poole2019variational}, and the lower bound is thus estimated by
\begin{align}
&{\rm div}_{\rm js}\bigl(p(\x^{(1)},\x^{(2)})\|p(\x^{(1)})p(\x^{(2)})\bigr)\notag\\
\geq&\mathbb E_{p(\x^{(1)},\x^{(2)})}\log\bigl( d(\x^{(1)},\x^{(2)}|\theta)\bigr)\notag\\
+&\mathbb E_{p(\x^{(1)})p(\x^{(2)})}  \log\bigl(1-d(\x^{(1)},\x^{(2)}|\theta)\bigr)\notag\\
=&{\rm ILBO}=-\mathcal L_{\rm{cl}}\label{jsd}
\end{align}
where ${\rm div}_{\rm js}$ denotes the Jensen-Shannon divergence.
\red{Compared with the Kullback-Leibler divergence, there is an advantage of the Jensen-Shannon divergence: the Jensen-Shannon divergence is symmetrical which can objectively measure the distance between two distributions to avoid different training results due to different sequences. }

We argue that the estimation of information entropy can be achieved by gradient descent over neural networks, \ie, we use $d(\x^{(1)}, \x^{(2)}|{\theta})$ as its neural network and define $\mathcal L_{\rm{cl}}$ to be its loss function. It can be seen from Eq.~\eqref{jsd} that maximizing ILBO is equivalent to minimizing the loss function $\mathcal L_{\rm{cl}}$. For a specific contrastive learning task, we can seek a parameterized discriminator $d(\x^{(1)}, \x^{(2)}|{\theta})$ to approach the lower bound of the mutual information ${\rm I}(X^{(1)}, X^{(2)})$. The process of maximizing ILBO defines an entropy neural estimation.

 By the aid of Eq.~\eqref{jsd}, we can exploit the contrastive loss, \ie, $\mathcal L_{\rm cl}$, to guide the discriminator $d(\x^{(1)},\x^{(2)}|\theta)$ in estimating the mutual information between two subsets, where $d(\x^{(1)},\x^{(2)}|\theta)$ is a discriminator with parameters $\theta$ that estimates the mutual information of the two subsets. That is, the discriminator $d(\x^{(1)},\x^{(2)}|\theta)$ maximizes the lower bound of the mutual information between subsets $X^{(1)}$ and $X^{(2)}$ under the supervision of $\mathcal  L_{\rm cl}$. This will be further discussed in $\S$\ref{Dis}.

In addition to the contrastive loss, the error between the two representations are also minimized.
Suppose $\z^{(i)} = f(\x^{(i)} | \theta), \forall\,i\in\{1,2\}$ denotes the output embedding, we hypothesize the error $\bm{\varepsilon}_i$ of cross-view embeddings is governed by the Gaussian distribution following Eq.~\eqref{eq-gaussian}, \ie,
\begin{align}\label{eq-gaussian}
\bm{\varepsilon}_i
\sim \frac{1}{\sqrt{2\pi}}\exp\Bigl(-\frac{\|\bm{\varepsilon}_i\|^2_2}{2}\Bigr)
\end{align}
Suppose our data point $\x\in X$ is i.i.d., the cross-view consistency, $\mathcal L_{\rm cvc}$, is defined by taking the negative logarithm of likelihood of the error,
\begin{equation}\label{8}
\mathcal L_{\rm cvc}
=\frac1n\sum_{i=1}^n{\left\|\bm{\varepsilon}_i\right\|}_2^2
=\frac1n\sum_{i=1}^n{\left\|\z_{i}^{(1)}-\z_{i}^{(2)}\right\|}_2^2\,.
\end{equation}
\red{The cross-view consistency is an optimal matching metric between two output embeddings which measures how similar they are. It means that the similar data points have similar embedding features. }

\red{Through the above analysis, we concentrate two different objectives, namely the contrastive loss and the cross-view consistency, into one loss function. Then, the overall objective function for the overall learning process is given by}
\begin{equation}\label{eq:overall_loss}
\mathcal L=\mathcal L_{\rm cl}+\lambda \mathcal L_{\rm cvc}
\end{equation}
where $\lambda$ is a hyper-parameter for balancing the contrastive term and the cross-view consistency term.

\subsection{Discriminator}\label{Dis}
In our framework, the neural discriminator estimates the mutual information between the two subsets sampling from the raw dataset. According to Eq.~\eqref{jsd}, minimizing the contrastive loss is equivalent to maximizing the mutual information between the two sampled subsets and forces the model to explore intrinsic structure determined by ${\rm E}(X)$. After multiple sampling, the distribution of the sampled subsets will approach the distribution of the raw dataset. That is, after multiple training with different samplings in each epoch, the discriminator approximately estimates the information entropy of the raw dataset, ${\rm E}(X)$.

In practice, we construct our discriminator by adapting two parameter-shared graph encoders $f$ to the Siamese networks~\cite{LeCun93NIPS,LeCun1467314,hadsell2006dimensionality}.
Given two views $X^{(1)}$ and $X^{(2)}$, we use the parameter-shared encoder $f$ to generate the pairwise representations of $\x_i^{(1)}$ and $\x_j^{(2)}$\,,
\begin{align}
\z_i^{(1)}&= f(\x_i^{(1)}|\theta), \x_i^{(1)}\in X^{(1)}\,,\\
\z_j^{(2)}&= f(\x_j^{(2)}|\theta), \x_j^{(2)}\in X^{(2)}
\end{align}
where $\theta$ is the parameter of the network $f$ and $z_i^{(1)}$ and $z_j^{(2)}$ are the latent embeddings corresponding to the input $\x_i^{(1)}$ and $\x_j^{(2)}$, respectively.
The output similarity score $d_{ij}$ is calculated by the inner product of the pairwise representations from different views,
\begin{align}
&D=\bigl(Z^{(1)}\bigr)^\T Z^{(2)}
=[d_{ij}(\x_i^{(1)},\x_j^{(2)}|\theta)]\notag\\
&=
\left[
  \begin{array}{cccc}
\red{(\z_1^{(1)})^\T\z_1^{(2)}} & \blue{(\z_1^{(1)})^\T\z_2^{(2)}} & \blue{\cdots} & \blue{(\z_1^{(1)})^\T\z_n^{(2)}} \\
\blue{(\z_2^{(1)})^\T\z_1^{(2)}} & \red{(\z_2^{(2)})^\T\z_2^{(2)}} & \blue{\cdots} & \blue{(\z_2^{(1)})^\T\z_n^{(2)}} \\
\blue{\vdots} & \blue{\vdots} & \red{\ddots} & \blue{\vdots} \\
\blue{(\z_n^{(1)})^\T\z_1^{(2)}} & \blue{(\z_n^{(1)})^\T\z_2^{(2)}} & \blue{\cdots} & \red{(\z_n^{(1)})^\T\z_n^{(2)}} \\
  \end{array}
\right]\,.
\end{align}
Our discriminator produces the similarity score for two arbitrary inputs.
Under the supervision of $\mathcal L_{\rm cl}$, we estimate the mutual information of $X^{(1)}$ and $X^{(2)}$ by maximizing the scores of positive data pairs from the joint distribution $p(\x_i^{(1)},\x_i^{(2)})$ while minimizing the scores of negative data pairs drawn independently from the marginal distribution $p(\x_i^{(1)})p(\x_j^{(2)})$~\cite{belghazi2018mutual,oord2018representation}. We use $\mathcal P$ and $\mathcal N$ to denote the sets of positive pairs and negative pairs, respectively.

With the help of the Monte-Carlo method, the contrastive objective of $\mathcal L_{\rm cl}$ can be explicitly expressed by
\begin{align}
\mathcal L_{\rm cl}
=&- \frac{1}{|\mathcal P|}\sum_{(i,i)\in\mathcal P} \log d_{ii}(\x_i^{(1)},\x_i^{(2)}|\theta)\notag\\
&-\frac{1}{|\mathcal N|}\sum_{{(i,j)\in\mathcal N}}  \log\bigl(1-d_{ij}(\x_i^{(1)},\x_j^{(2)}|\theta)\bigr)\label{loss_cl}
\end{align}
where $\x_i^{(1)}$ and $\x_i^{(2)}$ are belong to positive pairs $(i,i)\in\mathcal P$ while $\x_i^{(1)}$ and  $\x_j^{(2)}$ are negative pairs $(i,j)\in\mathcal N, \forall\,i\neq j$. \red{After the above information ensemble, the upper bound of the mutual information is the information entropy of the raw dataset. Furthermore, because of the input subset sampling and the construction of supervision pairs, our contrastive objective of $\mathcal L_{\rm cl}$ is more powerful and has better prediction performance, which is also empirically confirmed in $\S$\ref{Abl}. }

\subsection{Maximizing ILBO}
The process of maximizing ILBO is to learn the graph encoder $f$ parameterized by $\theta$ in the discriminator.
It includes two aspects during the training process, \ie, the input subset sampling and the construction of supervision pairs.
\subsubsection{Subset sampling}
Nodes and edges are two fundamental elements in a graph.
\red{Data sampling is based on bootstrapping, so using different subsets renders the graph encoder ensemble information of the raw set.}
The subset sampling is straightforward.
For each node feature $\bm h_{i}$ or each edge $a_{ij}$, we only have two choices: drop it or keep it.

The designed sampling approach naturally needs to select the input sets of both nodes and edges.
Generally, we follow very simple rules to sample them: 1) each node and edge are randomly selected \wrt the \textit{Bernoulli} distribution; 2) to maintain the structure of the input graph, values of the dropped elements are set to zeros.

Particularly,  the node feature selection follows Eq.~\eqref{eq:node_drop}
\begin{equation}\label{eq:node_drop}
\bm h_{i} = v\bm h_{i}, \,\,v\sim{\rm Bern}(v|1-p_h)
\end{equation}
where ${\rm Bern}(v|1-p_h)$ is the Bernoulli distribution with the drop rate $p_h$ and the value $v$ is 0 or 1.

Analogously, the edges can be selected in the same way, which is also inspired by DropEdge~\cite{rong2019dropedge,zhu2020deep}.
We consider a direct way for sampling input graphs where we randomly drop edges in the graph with the DropEdge rate $p_a$. After edge sampling, the new adjacency matrix $\hat A=[\hat a_{ij}]$ is given by,
\begin{equation}\label{aij}
\hat a_{ij}=
\begin{cases}
0,&a_{ij}=0\,;\\
u,&a_{ij}=1\, {\rm and}\, u\sim{\rm Bern}(u|1-p_a)
\end{cases}
\end{equation}
\red{where ${\rm Bern}(u|1-p_a)$ is the Bernoulli distribution, and $p_a < 1$.}
For the edge level sampling, $\hat a_{ij}$ is set to zero for the dropped edges with Eq.~\eqref{aij}.

\subsubsection{Construction of supervision pairs}
As illustrated in Eq.~\eqref{loss_cl}, the mutual information estimation between views/subsets turns to maximize scores of positive pairs while minimizing scores of negative pairs under the supervision of the contrastive loss. Therefore, it is critical to construct the supervision pairs carefully. In this section, we present a well-designed selection strategy of positive and negative pairs.

In each batch, we feed $X^{(1)}=[\x_1^{(1)},\x_2^{(1)},\ldots,\x_n^{(1)}]$ and $X^{(2)}=[\x_1^{(2)},\x_2^{(2)},\ldots,\x_n^{(2)}]$ into the graph encoder,
and obtain the corresponding latent embeddings $Z^{(1)}=[\z_1^{(1)},\z_2^{(1)},\ldots,\z_n^{(1)}]$ and $Z^{(2)}=[\z_1^{(2)},\z_2^{(2)},\ldots,\z_n^{(2)}]$ where $n$ is the batch size\,.
As the encoder network does not change the mapping structure of the graph, items have the same indices in $X$ and $Z$, \ie, since the encoder is a one-to-one mapping, shuffling $X$ has the same effect of shuffling $Z$.
If we shuffle the indices of the embedding $Z$, the correspondence between the nodes in the two different views,~\ie, $Z^{(1)}$ and $Z^{(2)}$, will be disrupted. In the classic strategy, positive pairs have the same indices from $Z^{(1)}$ and $Z^{(2)}$. Negative pairs are selected who have the same index from $Z^{(1)}$ and the shuffled $\hat Z^{(2)}$. Therefore, positive and negative pairs can be conveniently determined.

The two subsets can be seen as two views of the raw dataset, so the score $d_{ij}$ of positive and negative pairs can be selected in the cross-view similarity matrix $D$. In order to enlarge the sets $\mathcal P$ and $\mathcal N$, we obtain a cross-view similarity matrix $D$ by
\begin{align}\label{Dmatix}
D=[d_{ij}]
=\bigl(Z^{(1)}\bigr)^\T Z^{(2)}
=\bigl[(\z^{(1)}_i)^\T\z^{(2)}_j\bigr]
\end{align}
where $d_{ij}=(\z^{(1)}_i)^\T\z^{(2)}_j$ is an element of $D$\,. By using the similarity matrix $D=[d_{ij}]$,
we modify the contrastive loss to enhance the representation ability of the encoder.
As shown in Eq.~\eqref{Dmatix}, the positive pairs are diagonal elements definitely.
In the non-diagonal parts of $D$, high values indicate that the pairs have a larger probability of belonging to the same class, so we add them to the set $\mathcal P$ of positive pairs. Candidate negative pairs have lower scores in the non-diagonal parts. The classic positive pairs $d_{ii}$ only locate at the main diagonal of the cross-view similarity matrix, we further add more positive pairs in non-diagonal whose scores are high, and negative pairs are those who have low scores. By using the similarity matrix $D=[d_{ij}]$, we modify the contrastive loss to enhance the representation ability of the encoder.

In the non-diagonal parts of Eq.~\eqref{Dmatix}, we select the top $k$ largest pairs to add in $\mathcal P$ and find the top $l$ smallest pairs as the negative.
A high score of similarity $d_{ij}$ implies a high probability of belonging to the same category and the pair $(i,j)$ can be regarded as a positive pair.
Negative pairs have low scores. If node $1$ and node $2$ have a high similarity score of $(\z_1^{(1)})^\T\z_2^{(2)}$, we will add $(1,2)$ into set $\mathcal P$.
If the pair (1,2) actually belongs to the same category, it implies that such a selection strategy helps $\mathcal{M}$-ILBO to improve the representation ability.

With the training process, the information between nodes is gradually gathered and the dependency cues between nodes are progressively determined with the supervision between nodes.
The representation ability improves in the training procedure. In each batch, the contrastive loss guides the model in estimating the mutual information between the two subsets and updates the model. The upper bound of the mutual information of the subsets is the information entropy of the raw set since two subsets are from the same raw set with bootstrapping. Training with different subsets enables data distribution to cover the entire data $X$. The positive and negative sample selection strategy maximizes the mutual information in the direction of the ground truth.
\subsubsection{Algorithm}
We summarize the proposed $\mathcal{M}$-ILBO approach in Algorithm~\ref{algorithm}.
The input graph is denoted by $X=(H,A)$ where $H=[{\bm h}_i]$ is the node feature and $A=[a_{ij}]$ is the adjacency matrix.
Specifically, we first sample two subsets as views using the proposed bootstrapping-based graph subset sampling strategy. Then, we send the features of both views into a graph neural network with shared parameters. A well-designed contrastive loss is utilized to enhance the representation ability of the encoder.
\begin{algorithm}[h]
\caption{The $\mathcal{M}$-ILBO algorithm.}\label{algorithm}
\begin{algorithmic}[1]
\STATE \textbf{Input}: Graph data $X=(H, A)$, node feature drop rate $p_h$, edge drop rate $p_a$, and parameter $\lambda$\,.
\STATE \textbf{Output}: $Z$\,.
\STATE \textbf{Initialization}: $epoch=0$, $epoch_{\max}$, and model parameter $\theta$\,.
\WHILE{$epoch\leq {epoch_{\max}}$}
    \STATE Sample $X$ for subset 1: $X^{(1)}$\,;
    \STATE Sample $X$ for subset 2: $X^{(2)}$\,;
    \STATE $Z^{(1)}=f(X^{(1)}|\theta)$\,;
    \STATE $Z^{(2)}=f(X^{(2)}|\theta)$\,;
    \STATE $D=\bigl(Z^{(1)}\bigr)^\T Z^{(2)}$\,;
    \STATE Select $\mathcal P$ and $\mathcal N$ from $D=[d_{ij}]$\,;
    \STATE Compute the loss by $\mathcal L=\mathcal L_{\rm cl}+\lambda \mathcal L_{\rm cvc}$\,;
    \STATE Update parameter $\theta$ by back propagation\,;
    \STATE $epoch = epoch + 1$\,;
\ENDWHILE
\STATE \textbf{Output}: $Z=f(X|\theta^\star)$\,.
\end{algorithmic}
\end{algorithm}

Besides, a consistency constraint renders each view to produce common information. The cross-view consistency loss renders each view to obtain the common information. In order to learn the consensus information in both views, the encoder's parameters of both views are shared.

As shown in Algorithm~1, we propose a novel genre of graph contrastive learning that simulates view augmentation via subset sampling from raw data and contrastively maximizes mutual information across these subsets, rather than using augmented multi-view data. Our approach theoretically guarantees ILBO for a given dataset, as we demonstrate in our theoretical analysis. To estimate the entropy, we utilize the Jensen-Shannon divergence parameterized by a neural network, thereby maximizing ILBO. During training, we randomly sample a large number of node features and graph edges for bootstrapping, similar to the effect of Dropout. The graph encoder parameters then gradually learn to maximize ILBO for different subsets. Two parameter-shared graph encoders are used to generate high-dimensional representations of two subsets, and discrimination scores are calculated based on the similarity of node pairs from both subsets. These scores guide the selection of positive and negative pairs, enriching the diversity of constructed self-supervisions. The training process is thus transformed into learning correlations between two subsets of the given graph data. This process is illustrated in Fig.~1 and Algorithm~1.

Benefiting from both the contrastive loss and the cross-view consistency loss, $\mathcal{M}$-ILBO algorithm gathers more and more task-relevant information. With the contrastive loss, $\mathcal{M}$-ILBO gathers more and more task-relevant information and obtains better representations. 
\subsection{Discussion}
\subsubsection{$\mathcal{M}$-ILBO} Intuitionally, our $\mathcal{M}$-ILBO roots in two classical techniques, \ie,  Bagging~\cite{breiman1996bagging} and Dropout ~\cite{srivastava2014dropout}. Here, their connections are discussed and analyzed.

Connection to Bagging: $\mathcal{M}$-ILBO samples two different subsets as the two views, $X^{(1)}$ and $X^{(2)}$, to train the model in each epoch.
That is, the different inputs feed into one encoder in different epochs, so model parameters are updated with two input subsets in each epoch and the updated model is seen as the pre-trained model for the next epoch.

In each epoch, because of using different inputs, the model is trained in an ensemble way. Iteratively, one updated model is further updated as the pre-trained model in the next epoch. After multiple iterations, it is equivalent to training multiple models, and each model participates in an ensemble on the final result, which improves the generalization ability of the model. The optimized model can be seen as the fusion of multiple discriminators. With the trained model, the final representation is obtained from the raw dataset $X$. The process simulates classical Bagging that generates multiple classifiers by subsets of the training set.

Connection to Dropout: The effect of the information ensemble of our approach is somewhat similar to the effect of Dropout~\cite{srivastava2014dropout} in deep neural networks, which drops neurons in proportion in each batch. By using Dropout, a sub-network is trained in each batch, while our discriminator is trained with different input subsets.

The ensemble of multiple models generally yields better predictions than a single one.
Bagging~\cite{breiman1996bagging} has shown great potentiality in improving performance~\cite{bauer1999empirical}.
Bagging is a method for generating multiple classifiers by subsets of the training set. Then, these classifiers are combined by voting, which theoretically reduces the variance of predictions and improves performance~\cite{buhlmann2002analyzing}.

In this paper, we try to learn in the form of Bagging~\cite{breiman1996bagging} and Dropout~\cite{srivastava2014dropout}.
Referring to the Dropout effect in deep neural networks~\cite{srivastava2014dropout}, we utilize multiple subsets to train one parameter-shared encoder.

Our discriminator estimates the mutual information of the two subsets of the raw input.
In each epoch, the contrastive loss maximizes the mutual information of two subsets of the raw dataset. After multiple sampling, the distribution of the sampled subsets will cover the distribution of the raw dataset.
Thus, after multiple training with different inputs of each epoch, the discriminator estimates the information entropy of the raw dataset.
\subsubsection{Representation Ability} 

To improve the representation ability of $\mathcal{M}$-ILBO, we consider exploiting a graph encoder with a contrastive loss and a cross-view consistency learning loss. For the contrastive loss, a well-designed selection strategy of positive and negative pairs is proposed in this paper. The encoder and two loss functions are useful to improve the representation ability.

A good view encoder $f$ is able to aggregate useful information while removing redundant information~\cite{tishby2000information,wu2020graph}, \ie, $\min\, {\rm I}(X,Z)-\beta{\rm I}(Y,Z)$ where we suppose that $Y$ is the ground-truth label indication matrix denoting the target of the downstream task, and $\beta$ is the trade-off hyper-parameter. If there is an ideal encoder, it will compress the entropy of $X$ to the entropy of $Y$. We use graph convolutional network (GCN)~\cite{kipf2016semi} as the graph encoder since GCN has powerful graph data encoding and compression ability. At the beginning, its representation ability is low, so the entropy of $Z$ is far from $Y$. If the encoder is trained very well, it will tend to close to $Y$.

With the contrastive loss, the cross-view consistency loss and a well-designed positive and negative pairs selection strategy, $\mathcal M$-ILBO gathers more and more information from task-relevant information and obtains better representations. Fig.~\ref{scheme} shows that the representation ability improves with the increase of the epochs. In each epoch, the contrastive loss guides the encoder in estimating the mutual information between the two subsets. The upper bound of their mutual information is the information entropy of the raw set since two subsets are from the same raw set. Training with different subsets enables features to cover the information of the entire data. The positive and negative sample multiple selection strategy maximizes the mutual information in the direction of ground-truth indication matrix $Y$.

\section{Experiments}
We conduct various experiments to evaluate the proposed $\mathcal{M}$-ILBO method on node classification tasks.
The focuses of the experiments are to validate the representation ability of the learned features, and the effectiveness of different contrastive objectives.
We report the classification \textit{accuracy} which is the percentage of nodes being correctly predicted.

We use different datasets for experiments. We compare our $\mathcal{M}$-ILBO approach to different baselines on the citation network datasets and the co-occurrence network datasets.

\red{The three citation networks are Cora~\cite{mccallum2000automating}, Citeseer~\cite{giles1998citeseer}, and Pubmed~\cite{namata2012query}.
Specifically, in these datasets, nodes represent papers, and edges denote the citation relationship. The bag-of-words representation of papers are regarded as node features, and labels are academic fields.}

\red{The four co-occurrence networks are Computer~\cite{shchur2018pitfalls}, Photo~\cite{shchur2018pitfalls}, CS~\cite{shchur2018pitfalls}, and Physics~\cite{shchur2018pitfalls}. Specifically, computer~\cite{shchur2018pitfalls} and Photo~\cite{shchur2018pitfalls} are two co-purchase graphs from Amazon, which nodes are products and the edge between two nodes indicates that two products are purchased at the same time. The sparse bag-of-words attribute vector is node feature. CS~\cite{shchur2018pitfalls} and Physics~\cite{shchur2018pitfalls} are two academic networks of co-authorship relationship from Microsoft Academic Graph. Nodes denote authors, and edges denote co-author relationships. The bag-of-words representation of the paper keywords are regarded as node features, and labels are academic fields.}

In order to highlight the outstanding representation learning performance of $\mathcal{M}$-ILBO, we compare the proposed $\mathcal{M}$-ILBO method with the unsupervised methods and the supervised methods.
Specifically, the unsupervised methods include DeepWalk~\cite{perozzi2014deepwalk}, GAE~\cite{kipf2016variational}, DGI~\cite{velickovic2019deep}, MVGRL~\cite{hassani2020contrastive}, GRACE~\cite{zhu2020deep}, GCA~\cite{zhu2021graph}, and CCA-SSG~\cite{zhang2021canonical}; and the supervised methods include MLP, Label Propagation (LP)~\cite{zhu2003semi}, GCN~\cite{kipf2016semi}, and GAT~\cite{velivckovic2017graph}.

We first train the model on all the nodes in a graph following Algorithm~\ref{algorithm}. Following the setting in DGI~\cite{velickovic2019deep}, after the encoder is trained, we test $X$ by feeding it into the learned graph encoder and obtain $Z$. $Z$ is fed into a downstream classifier,~\ie, linear logistic regression, to output the predicted labels for each node. For the logistic regression classifier, we only use the training set for optimizing the classifier and output the classification accuracy on the testing set. Following the setting in DGI~\cite{velickovic2019deep}, The classification accuracy in Tables~\ref{node1} and \ref{node2} are obtained by the logistic regression classifier.
The drop rate $p_h$ and $p_a$ are selected in the range of $[0,0.99]$. $\lambda$ is selected in the range of $[0.05,0.4]$.
We train $\mathcal{M}$-ILBO using Adam optimizer~\cite{kingma2014adam} where the learning rate is set to be $10^{-3}$. 
Without loss of generality, we use the GCN as the backbone, and the number of network layers is two.

\begin{table*}[t]
  \centering
  \caption{Accuracy (mean$\%\pm$std$\%$) comparison on citation network datasets. ($H$ denotes node features, $A$ is the adjacency matrix, $S$ is the diffusion matrix, and $Y$ is the label indication matrix).
  The proposed method outperforms the baselines.
    The best values are in \textbf{bold}, while the second-best numbers are underlined.
  }
  \label{node1}
  \centering
  \begin{tabular*}{0.94\textwidth}{@{\extracolsep{\fill}\,}lllccc}
  \toprule
  &Methods&Input&Cora&Citeseer&Pubmed\\
  \midrule  
  \multirow{4}*{Supervised}
  &MLP~\cite{velivckovic2017graph}  &$H,Y$    &55.1&46.5&71.4\\
  &LP~\cite{zhu2003semi}            &$A,Y$    &68.0&45.3&63.0\\
  &GCN~\cite{kipf2016semi}          &$H,A,Y$  &81.5&70.3&79.0\\
  &GAT~\cite{velivckovic2017graph}  &$H,A,Y$  &83.0$\pm$0.7&72.5$\pm$0.7&79.0$\pm$0.3\\
  \midrule  

  \multirow{9}*{Unsupervised}
  &Raw Features~\cite{velickovic2019deep}   &$H$    &47.9$\pm$0.4&49.3$\pm$0.2&69.1$\pm$0.3\\
  &DeepWalk ~\cite{perozzi2014deepwalk}  &$H,A$    &70.7$\pm$0.6&51.4$\pm$0.5&74.3$\pm$0.9\\
  &GAE~\cite{kipf2016variational}        &$H,A$     &71.5$\pm$0.4&65.8$\pm$0.4&72.1$\pm$0.5\\
  &DGI~\cite{velickovic2019deep}         &$H,A$     &82.3$\pm$0.6&71.8$\pm$0.7&76.8$\pm$0.6\\
  &MVGRL~\cite{hassani2020contrastive}   &$H,A,S$   &83.5$\pm$0.4&\underline{73.3$\pm$0.5}&80.1$\pm$0.7\\
  &GRACE~\cite{zhu2020deep}              &$H,A$     &81.9$\pm$0.4&71.2$\pm$0.5&80.6$\pm$0.4\\
  &CCA-SSG~\cite{zhang2021canonical}     &$H,A$     &\underline{84.2$\pm$0.4}&73.1$\pm$0.3&\underline{81.6$\pm$0.4}\\
  &\textbf{$\bm{\mathcal{M}}$-ILBO}      &$H,A$     &\textbf{85.7$\pm$0.3} & \textbf{74.2$\pm$0.7} & \textbf{81.8$\pm$0.3}\\
    \bottomrule
    \end{tabular*}
\end{table*}

\begin{table*}[t]
    \centering
    \caption{Accuracy (mean\%$\pm$std\%) comparison on co-occurrence network datasets.
    The proposed method achieves competitive performance compared to the baselines.
    The best values are in \textbf{bold}, while the second-best numbers are underlined.
    }
    \label{node2}
    \centering
    \begin{tabular*}{0.988\textwidth}{@{\extracolsep{\fill}\,}llcccc}
      \toprule
    Methods&Input&Computer&Photo&CS&Physics\\
    \midrule
	Supervised GCN~\cite{kipf2016semi}          &$H,A,Y$       &86.51$\pm$0.54&92.42$\pm$0.22&93.03$\pm$0.31&95.65$\pm$0.16\\
  Supervised GAT~\cite{velivckovic2017graph}  &$H,A,Y$      &86.93$\pm$0.29&92.56$\pm$0.35&92.31$\pm$0.24&95.47$\pm$0.15\\
  \midrule
	Raw Features~\cite{velickovic2019deep}   &$H$        &73.81$\pm$0.00&78.53$\pm$0.00&90.37$\pm$0.00&93.58$\pm$0.00\\
  DeepWalk~\cite{perozzi2014deepwalk}      &$H,A$      &86.28$\pm$0.07&90.05$\pm$0.08&87.70$\pm$0.04&94.90$\pm$0.09\\
  GAE~\cite{kipf2016variational}           &$H,A$      &85.27$\pm$0.19&91.62$\pm$0.13&90.01$\pm$0.71&94.92$\pm$0.07\\
  DGI~\cite{velickovic2019deep}            &$H,A$      &83.95$\pm$0.47&91.61$\pm$0.22&92.15$\pm$0.63&94.51$\pm$0.52\\
  MVGRL~\cite{hassani2020contrastive}      &$H,A,S$    &87.52$\pm$0.11&91.74$\pm$0.07&92.11$\pm$0.12&95.33$\pm$0.03\\
  GRACE~\cite{zhu2020deep}                 &$H,A$      &86.25$\pm$0.25&92.15$\pm$0.24&92.93$\pm$0.01&95.26$\pm$0.02\\
  GCA~\cite{zhu2021graph}                  &$H,A$      &87.85$\pm$0.31&92.49$\pm$0.09&93.10$\pm$0.01&\textbf{95.68$\pm$0.05}\\
  CCA-SSG~\cite{zhang2021canonical}        &$H,A$      &\underline{88.74$\pm$0.28}&\underline{93.14$\pm$0.14}&\textbf{93.31$\pm$0.22}&95.38$\pm$0.06\\
  \textbf{$\bm{\mathcal{M}}$-ILBO}         &$H,A$      & \textbf{89.16$\pm$0.09} & \textbf{93.73$\pm$0.27} & \underline{93.23$\pm$0.09} & \underline{95.43$\pm$0.09}\\
    \bottomrule
    \end{tabular*}
\end{table*}

\subsection{Comparison with the state-of-the-arts}
In Tables \ref{node1} and \ref{node2}, we separately report the performance of the proposed method compared to a variety of state-of-the-art node classification results on the citation network datasets and the co-occurrence network datasets.

We have the following observations:
  (1) Among all self-supervised learning strategies, $\mathcal{M}$-ILBO achieves competitive performance on seven datasets, showing state-of-the-art results on most datasets. To be specific, previous works such as DeepWalk~\cite{perozzi2014deepwalk} and DGI~\cite{velickovic2019deep} merely leverage a single raw dataset. In contrast, $\mathcal{M}$-ILBO is trained with an information ensemble, and it successfully leverages the ensemble of multiple subsets of the raw dataset to yield better predictions.
  (2) It is easy to find that the reconstruction-based method,~\eg, GAE, has lower accuracy than $\mathcal{M}$-ILBO.
$\mathcal{M}$-ILBO emphasizes that the model generates representations that include useful information by contrastive learning.
  (3) Without the supervision of the ground-truth labels, our results are higher than that of supervised methods. $\mathcal{M}$-ILBO assembles rich predictive information from the raw dataset \red{so that the learned representations are more robust and expressivity, leading to better performance on node classification task}.
Training with different subsets enables features to cover the information of the entire data.
The positive and negative sample multiple selection strategy maximizes the mutual information in the direction of task correlation, which contributes to the efficiency of $\mathcal{M}$-ILBO.


\begin{table}[t]
  \caption{Ablation study of node classification. The best values are in \textbf{bold}.}\label{ablation}
  \centering 
\begin{tabular*}{0.48\textwidth}{@{\extracolsep{\fill}\,}llccc}
\toprule
Methods &Input &Cora &Citeseer &Pubmed\\
\midrule
only consistency                   &$H,A$          &82.0    &73.2   &70.1 \\
shuffling strategy                 &$H,A$          &83.9    &72.9   &75.8 \\
\textbf{$\bm{\mathcal{M}}$-ILBO}        &$H,A$     &\textbf{85.5}    &\textbf{73.7}   &\textbf{82.2} \\
\bottomrule
  \end{tabular*}
\end{table}

\subsection{Ablation study}\label{Abl}
In order to demonstrate the merits of the strategy of supervision pairs, we compare it with the classic shuffling strategy. We also leave out the contrastive loss and only use the consistency loss to show that learning is useful for improving the representation ability. We name the three experiments as ``$\mathcal{M}$-ILBO strategy'', ``shuffling strategy'', and ``only consistency''. We conduct the ablation study on citation datasets and the accuracy is shown in Tables~\ref{ablation}.

From Tables~\ref{ablation}, we have three observations:
(1) The accuracy metrics of  "$\mathcal{M}$-ILBO strategy'' are higher than the others, indicating the effectiveness and necessity of the modified contrastive loss.
We believe that this is an indication that $\mathcal{M}$-ILBO comprehensively explores the underlying semantic information and important connective structural information,
rather than just using samples with the same index in different subsets as positive pairs.
(2) The accuracy of ``shuffling strategy'' is $83.9\%$ on Cora dataset, while the result of ``$\mathcal{M}$-ILBO strategy'' has $1.6\%$ performance gain. This gap indicates a huge benefit from selection strategies of positive pairs and negative pairs.
(3) The accuracy metrics of ``only consistency'' are lower than the other two on Cora and Pubmed, which indicates that it is crucial to contrast negative pairs.
Because of the consistency loss, the distributions of the learned two-view representations are close to each other and representations tend to censuses semantic information.

The representation ability of the discriminator is boosted by the modified contrastive loss, and it does learn more discriminative representations than ``shuffling strategy'' and ``only consistency''.
\subsection{Parameter sensitivity analysis}
\begin{figure*}[!ht]
\centering
  \includegraphics[width=1\textwidth]{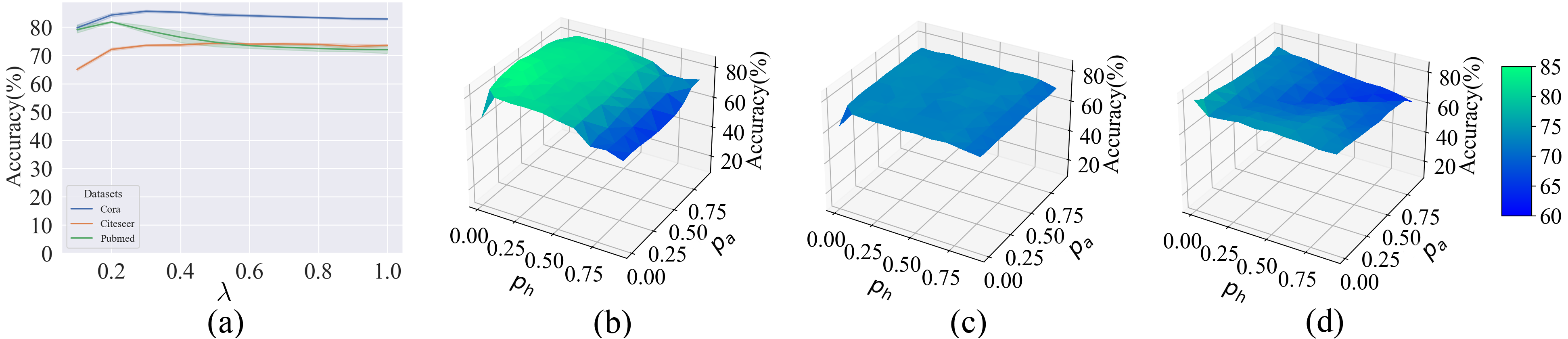}
  \caption{Parameter sensitivity experiments for $\lambda$, $p_h$ and $p_a$.}
  \label{parameters}
\end{figure*}

For unsupervised contrastive learning methods, parameter insensitivity is vital for enhancing their stability.
Since the value of $\lambda$ is an important parameter in the experiment, we design a sensitivity experiment of $\lambda$ on citation network datasets as shown in Fig.~\ref{parameters}(a). It can be seen from Fig.~\ref{parameters}(a) that the test accuracy is stable when its value varies in a range of [0.1, 1] with 0.1 intervals.

Figs.\,\ref{parameters}(b)-(d) show the sensitivity of $p_h$ and  $p_a$ on citation datasets.
Specifically, we set $p_h$ and $p_a$ to $\{0.01,0.1,\ldots,0.9\}$, and we report test accuracy. It can be
observed from Figs.\,\ref{parameters}(b)-(d) that competitive accuracy is
obtained over a wide range of $p_h$ and $p_a$\,.
\section{Related Work}
A typical contrastive learning model usually augments a raw dataset to construct multiple views and contrasts pairwise representations from different views~\cite{oord2018representation}. Contrastive learning aims to maximize the similarity scores of positive pairs and minimize that of negative pairs.
\red{Since maximizing the mutual information between different representations is theoretical basis of contrastive learning, InfoNCE estimates the mutual information by maximizing its lower bound~\cite{oord2018representation}.}
The contrastive loss can be given by Eq.~\eqref{loss_cl}. Suppose that positive pairs are from joint distribution and negative pairs are from the product of marginal distributions, so  Eq.~\eqref{loss_cl} effectively maximizes the mutual information between $X^{(1)}$ and $X^{(2)}$~\cite{oord2018representation}.

In graph contrastive learning, Deep Graph InfoMax (DGI)~\cite{velickovic2019deep} maximizes the mutual information between a global feature and a local feature. DGI is inspired by the previous success of the Deep InfoMax (DIM)~\cite{hjelm2018learning}. DIM maximizes the mutual information between the image patch and corresponding learned high-level representations.

MVGRL~\cite{hassani2020contrastive} expands DGI and exploits two graphs. Both DGI and MVGRL maximize the mutual information between the local feature and the global feature. The global feature is aggregated from the all node features of the whole graph, which is different from DIM. In DIM, a single instance is an image while it is a node in DGI. An image, its patch, and its representation have the same semantic information, and DIM is used for image classification. However, the node and the whole graph do not have similar semantic information in DGI. Later, some graph contrastive learning methods~\cite{peng2020graph,zhu2020deep,CG3} contrast node representations from different views.

In this paper, we concentrate on contrastive learning and maximize the agreement of different views. Most existing methods focus on feature information while ignoring the importance of category information, and we select positive pairs by further selecting samples of the same class in addition to different representations from the same instance.

Consistency regularization follows the continuity assumption~\cite{NIPS2014_66be31e4} argue that similar data points have basically consistent output after mapping~\cite{tarvainen2017mean}.
In other words, for unlabeled data, either perturb the model or data, the corresponding predicted results should be consistent.

Following this assumption, many literatures incorporate consistency in their methods. The most widely used and most basic consistency loss is the $\ell_2$ regularization. Ladder network~\cite{NIPS2015_378a063b} evaluates each data point with and without noise, and then applies a consistency cost between the two predictions to learn the denoising function, which maps the corrupted output onto the denoising targets.
Mean Teacher~\cite{tarvainen2017mean} defines the consistency cost as the expected distance between the prediction of the student model and the prediction of the teacher model.
$\Pi$-model~\cite{ICLR2017Temporal} encourages consistent network output under two different dropout conditions.

In the field of graph contrastive learning,
\red{the most of methods~\cite{peng2020graph,hassani2020contrastive,zhu2020deep,you2020graph} generate different views via perturbating graph structures. The intuition is that the data augmentation schemes force the model to learn insensitive representations of perturbations at unimportant nodes and edges. }
 GRACE~\cite{zhu2020deep} and GraphCL~\cite{you2020graph} generate views with different types of graph augmentations and learn node representations by maximizing the agreement of node representations in these two views.
Inherited from consistency theory, we maximize the agreement between the outputs of two subsets of input.

\section{Conclusion}
We introduce a concept of ILBO. ILBO and ELBO (the Evidence Lower BOund) are determined by the raw dataset and are maximized to find model weights. Dataset structure is determined by the entropy of the dataset and we introduced an entropy neural estimation by maximizing ILBO ($\mathcal{M}$-ILBO) algorithm for unsupervised contrastive learning.
Contrastive learning has demonstrated the success of the neural estimation of the mutual information between high-dimensional data. Contrastive learning algorithms maximize the mutual information of the two-view features. We use a novel contrastive loss to train a parameter-shared graph encoder by two sampled views of the raw dataset. Besides, a cross-view consistency learning loss renders each view to obtain the common information. 
We used the contrastive loss to train a parameter-shared graph encoder by two sampled subsets of the raw dataset. We sampled two subsets in every epoch and maximized ILBO. In each epoch of the training process, the contrastive loss guides the encoder in estimating the mutual information between the two subsets. The upper bound of their mutual information is the entropy of the raw dataset since the two subsets are sampled from the same raw dataset.

The method of estimating information entropy is to train with the subsets of the raw dataset.
The abundant experimental results indicate that the proposed $\mathcal{M}$-ILBO achieves superior performance on various datasets for graph representation learning compared to state-of-the-art methods.
\red{As mentioned above, constructing the supervision pairs has higher memory requirements than shuffling strategy, which can restrict its applicability for large datasets. This aspect is time-consuming and deserves improvement. }  
Extensive experiments under various conditions verify the performance of $\mathcal{M}$-ILBO in several graph datasets. 
Comparing to existing state-of-the-art unsupervised representation learning algorithms, $\mathcal{M}$-ILBO achieves impressive performances and exhibits satisfactory generalization ability.


\bibliographystyle{ACM-Reference-Format}
\bibliography{ID5293}


\end{document}